\documentclass[11pt,a4paper]{article}
\usepackage[hyperref]{acl2020}
\usepackage{times}
\usepackage{latexsym}
\usepackage{url}
\usepackage{booktabs}
\usepackage{adjustbox}
\usepackage{xcolor}
\usepackage{multirow}
\usepackage{amsmath,amssymb}

\usepackage{algorithm}
\usepackage{algorithmic}

\DeclareMathOperator*{\argmax}{arg\,max}

\newcommand{\pluseq}{\mathrel{+}=}

\newtheorem{definition}{Definition}

\aclfinalcopy 

\newcommand\DeEn{De$\Rightarrow$En }
\newcommand\ZhEn{Zh$\Rightarrow$En }
\newcommand\FrEn{Fr$\Rightarrow$En }
\usepackage{color}
\definecolor{mypink2}{RGB}{219, 48, 122}
\definecolor{jrcolor}{RGB}{100, 150, 225}
\definecolor{jrcomment}{RGB}{70, 200, 150}

\title{Evaluating Explanation Methods for Neural Machine Translation}

\author{Jierui Li$^{1}$~~~Lemao Liu$^{2}$\thanks{~~This work was done during J.Li \& G.Li's internship at Tencent AI Lab. L.Liu is the corresponding author.}~~~Huayang Li$^{2}$~~~Guanlin Li$^{3}$~~~Guoping Huang$^{2}$~~~Shuming Shi$^{2}$ \\
        $^{1}$University of Electronic Science and Technology of China \\
        $^{2}$Tencent AI Lab,~~
        $^{3}$Harbin Institute of Technology\\
        \tt \{lijierui19,epsilonlee.green\}@gmail.com, \\
        \tt \{redmondliu,alanili,donkeyhuang,shumingshi\}@tencent.com 
}

\begin{document}
\maketitle
\begin{abstract}

Recently many efforts have been devoted to interpreting the black-box NMT models, but little progress has been made on metrics to evaluate explanation methods.
Word Alignment Error Rate can be used as such a metric that matches human understanding, however, it can not measure explanation methods on those target words that are not aligned to any source word. This paper thereby makes an initial attempt to evaluate explanation methods from an alternative viewpoint. 
To this end, it proposes a principled metric based on \textit{fidelity} in regard to the predictive behavior of the NMT model. 
As the exact computation for this metric is intractable, we employ an efficient approach as its approximation. 
On six standard translation tasks, we quantitatively evaluate several explanation methods in terms of the proposed metric and we reveal some valuable findings for these explanation methods in our experiments.
\end{abstract}

\section{Introduction}

Neural machine translation (NMT) has witnessed great success during recent years~\cite{sutskever2014sequence,bahdanau2014neural,gehring2017convolutional,vaswani2017attention}.
One of the main reasons is that neural networks possess the powerful ability
to model sufficient context by entangling all source words and target words from translation history. The downside yet is its poor interpretability: it is unclear which specific words from the entangled context are crucial for NMT to make a translation decision. As interpretability is important for understanding and debugging the translation process and particularly to further improve NMT models, many efforts have been devoted to explanation methods for NMT~\cite{ding2017visualizing,alvarez2017causal,li-etal-2019-word,koehn2019saliency,he-etal-2019-towards}.
However, little progress has been made on evaluation metric to study how good these explanation methods are and which method is better than others for NMT.

Generally speaking, we recognize two orthogonal
dimensions for evaluating the explanation methods:
i) how much the pattern (such as source words) extracted by an explanation method matches \textit{human understanding} on predicting a target word;
or ii) how the pattern matches \textit{predictive behavior} of the NMT model on predicting a target word.
In terms of i), Word Alignment Error Rate (AER) can be used as
a metric to evaluate an explanation method by measuring
agreement between \textit{human-annotated}
word alignment and that derived from the explanation method.
However, AER can not measure explanation methods on those 
target words that are not aligned to any source words according to human annotation.

In this paper, we thereby make an initial attempt to measure explanation methods for NMT according to the second dimension of interpretability, which covers all target words.
The key to our approach can be highlighted as \textit{fidelity}: when extracting the most relevant words with an explanation method, if those relevant words have the potential to construct an optimal proxy model that agrees well with the NMT model on making a translation decision, then this explanation method is good (\S3).
To this end, we formalize a principled evaluation metric as an optimization problem over the expected disagreement between the optimal proxy model and the NMT model(\S3.1). Since it is intractable to exactly calculate the principled metric for a given explanation method, we propose an approximate metric to address the optimization problem. Specifically, inspired by statistical learning theory~\cite{vapnik1999overview}, we cast the optimization problem into a standard machine learning problem which is addressed in a two-step strategy:
firstly we follow empirical risk minimization to optimize the empirical risk; then we validate the optimized parameters on a held-out test dataset. Moreover, we construct different proxy model architectures by utilizing the most relevant words to make a translation decision, leading to variant approximate metric in implementation (\S3.2).

We apply the approximate metric to evaluate four explanation methods including attention~\cite{bahdanau2014neural,vaswani2017attention}, gradient norm~\cite{li2015visualizing}, weighted gradient~\cite{koehn2019saliency} and prediction difference~\cite{li-etal-2019-word}.
We conduct extensive experiments on three standard translation tasks for two popular translation models in terms of the proposed evaluation metric.
Our experiments reveal valuable findings for these explanation methods:
    1) The evaluation methods (gradient norm and prediction difference) are good to interpret the behavior of NMT;
    2) The prediction difference performs better than other methods.

This paper makes the following contributions:
\begin{itemize}
\item It presents an attempt at evaluating the explanation methods for neural machine translation from a new viewpoint of fidelity.
\item It proposes a principled metric for evaluation, and to put it into practice it derives a simple yet efficient approach to approximately calculate the metric.
\item It quantitatively compares several different explanation methods and evaluates their effects in terms of the proposed metric.
\end{itemize}

\section{NMT and Explanation Methods}

\subsection{NMT Models}
Suppose $\boldsymbol{x} = \lbrace x_1, \cdots, x_{|\boldsymbol{x}|} \rbrace$ denotes a source sentence with length $|\boldsymbol{x}|$ and $\boldsymbol{y} = \lbrace y_1, \cdots, y_{|\boldsymbol{y}|} \rbrace$ is a target sentence. 
Most NMT literature models the following conditional probability $P(\boldsymbol{y}\mid \boldsymbol{x})$ in an encoder-decoder fashion:
\begin{equation}\begin{array}{rcl}
    P\left(\boldsymbol{y}\mid \boldsymbol{x}\right)
    & = & \prod\limits _{t} P\left(y_t \mid \boldsymbol{y}_{<t}, \boldsymbol{x}\right) \\
    & = & \prod\limits _{t} P\left(y_t \mid s_t \right) ,
    \label{eq:nmt}
\end{array}\end{equation}
where $\boldsymbol{y}_{<t} = \lbrace y_1, \cdots, y_{t-1} \rbrace$ denotes a
prefix of $\boldsymbol{y}$ with length $t-1$, and $s_t$ is the decoding state vector of timestep $t$.   
In the encoding stage, the encoder of a NMT model transforms the source sentence $\boldsymbol{x}$ into a sequence of hidden vectors $\boldsymbol{h} =\lbrace h_1, \cdots, h_{|\boldsymbol{x}|} \rbrace$. In the decoding stage, the decoder module summarizes the hidden vectors $\boldsymbol{h}$ and the history decoding states $\boldsymbol{s}_{<t}= \lbrace s_1, \cdots, s_{t-1} \rbrace$ into the decoding state vector $s_t$. 
In this paper, we consider two popular 
NMT translation architectures, {\textsc{Rnn-Search}}~\cite{bahdanau2014neural} and {\textsc{Transformer}}~\cite{vaswani2017attention}. \textsc{Rnn-Search} utilizes a bidirectional RNN to define $\boldsymbol{h}$ and it computes $s_t$ by the attention function over $\boldsymbol{h}$, i.e., 
\begin{equation}
s_t=\mathrm{Attn}(s_{t-1}, \boldsymbol{h}),
\label{eq:rnn}
\end{equation}
where $\mathrm{Attn}$ is the attention function, which is defined as follows:
\begin{align}
\mathrm{Attn}(q, \boldsymbol{v})  &= \sum_i \alpha(q,v_i) v_i, \notag \\ 
\alpha(q, v_i) &= \frac{\exp\big(e(q, v_i)\big)}{\sum_j \exp\big(e(q, v_j)\big)} \label{eq:att},
\end{align}
\noindent where $q$ and $v_i$ are vectors, $e$ is a similarity function over a pair of vectors and $\alpha$ is its normalized function. 

Different from \textsc{Rnn-Search}, which relies on \textsc{Rnn}, \textsc{Transformer} employs an attention network to define $\boldsymbol{h}$, and two additional attention networks to define $s_t$ as follows:~\footnote{Due to space limitation, we present the notations for a single layer NMT models, and for \textsc{Transformer} we only keep the attention (with a single head) block while skipping other blocks such as resNet and layer normalization. More details can be found in the references~\cite{vaswani2017attention}. }
\begin{align}
\begin{split}
s_t  &= \mathrm{Attn}(s_{t+\frac{1}{2}}, \boldsymbol{h}), \\
s_{t+\frac{1}{2}} &= \mathrm{Attn}(s_{t-1}, \boldsymbol{s}_{<t}).
\end{split}
\label{eq:tfm}
\end{align}

\subsection{Explanation Methods}
In this section, we describe several popular explanation methods that will be evaluated with our proposed metric. Suppose $c_t=\langle \boldsymbol{y}_{<t}, \boldsymbol{x} \rangle$ denotes the context at timestep $t$, $w$ (or $w'$) denotes either a source or a target word in the context $c_t$. 
According to \citet{poerner-etal-2018-evaluating}, each explanation method for NMT could be regarded as a word relevance score function $\phi(w; y, c_t)$, where $\phi(w; y, c_t)>\phi(w'; y, c_t)$ indicates that $w$ is more useful for the translation decision $P(y_t|c_t)$ than word $w'$.

\paragraph{Attention}
Since ~\citet{bahdanau2014neural} propose the attention mechanism for NMT, it has been the most popular explanation method for NMT~\cite{tu-etal-2016-modeling,mi-etal-2016-supervised,liu-etal-2016-neural,zenkel2019adding}. 

To interpret \textsc{Rnn-Search} and \textsc{Transformer}, we define different $\phi$ for them based on attention. For \textsc{Rnn-Search}, since attention is only defined on source side, $\phi(w; y, c_t)$ can be defined only for the source words: $$\phi(x_i; y, c_t) = \alpha(s_{t-1}, h_i)$$
\noindent where $\alpha$ is the attention weight defined in Eq.\eqref{eq:att}, and $s_{t-1}$ is the decoding state of \textsc{Rnn-Search} defined in Eq.\eqref{eq:rnn}.
In contrast, \textsc{Transformer} defines the attention on both sides and thus $\phi(w; y, c_t)$ is not constrained to source words:
 
\begin{equation*}
\phi(w; y, c_t) = \left\{\begin{matrix}
 \alpha(s_{t+\frac{1}{2}}, h_i) & \textrm{ if $w=x_i$,} \\ 
\alpha(s_{t-1}, s_j)  &  \textrm{ if $w=y_j$ and $j<t$,} 
\end{matrix}\right.
\end{equation*}
where $s_{t-1}$ and $s_{t+\frac{1}{2}}$ are defined in Eq.\eqref{eq:tfm}.

\paragraph{Gradient}
Different from attention that is restricted to a specific family of networks,
the explanation methods based on gradient are more general. 
Suppose $g(w, y)$ denotes the gradient of $P(y \mid c_t)$ w.r.t to the variable $w$ in $c_t$:
\begin{equation}
g(w,y) = \frac{\partial P(y\mid c_t)}{\partial w}
\end{equation}
\noindent where $\partial w$ denotes the gradient w.r.t the embedding of the word $w$, since a word itself is discrete and can not be taken gradient. Therefore, $g(w,y)$ returns a vector with the same shape as the embedding of $w$.  
In this paper, we implement two different gradient-based explanation methods and derive different definitions of $\phi(w; y, c_t)$ as follows.

\begin{itemize}
\item {\bf Gradient Norm}~\cite{li2015visualizing}:
The first definition of $\phi$ is the $\ell-1$ norm of $g$:
$$\phi(w; y, c_t) = |g(w,y)|_{\ell-1}.$$
\item {\bf Weighted Gradient}~\cite{koehn2019saliency}:
The second one is defined as the weighted sum of the embedding of $w$, with the return of $g$ as the weight:
$$\phi(w; y, c_t) = g(w,y)^\top\cdot w.$$
\end{itemize}

It is worth noting that for each sentence $\langle \boldsymbol{x}, \boldsymbol{y} \rangle$, one has to independently calculate $\frac{\partial P(y\mid c_t)}{\partial w}$ for each timestep $t$. Therefore, one has to calculate $|\boldsymbol{y}|$ times of gradient for each sentence. In contrast, when training NMT, one only requires calculating sentence level gradient and it only calculates one gradient thanks to gradient accumulation in back propagation algorithm. 
\paragraph{Prediction Difference}
\citet{li-etal-2019-word} propose a prediction difference (\textsc{Pd}) method, which defines the contribution of the word $w$ by evaluating the change in the probability after removing $w$ from $c_t$.  
Formally, $\phi(w;y, c_t)$ based on prediction difference is defined as follows:
\begin{equation*}
\phi(w; y, c_t) = P(y\mid c_t) - P(y\mid c_t\backslash w)
\end{equation*}
\noindent where $P(y\mid c_t)$ is the NMT probability of $y$ defined in Eq.\eqref{eq:nmt}, and $P(y\mid c_t\backslash w)$ denotes
the NMT probability of $y$ after excluding $w$ from its context $c_t$. To achieve the effect of excluding $w$ from $c_t$, it simply replaces the word embedding of $w$ with zero vector before feeding it into the NMT model. 
 

\section{Evaluation Methodology}

\subsection{Principled Metric}

The key to our metric is described as follow: {to define an explanation method $\phi$ good enough in terms of our metric, the relevant words selected by $\phi$ from the context $c_t$ should have} the potential to construct an optimal model that exhibits similar behavior to the target model $P(y \mid c_t )$. To formalize this metric, we first {specify} some necessary notations.

Assume that $f(c_t)$ is the target word predicted by $P(y\mid c_t)$, i.e., $f(c_t)=\argmax_y P(y\mid c_t)$. In addition, let $\mathcal{W}_{\phi}^k(c_t)$ be the \textrm{top}-$k$ relevant words on the source side and target side of the context $c_t$:
\begin{multline*}
\mathcal{W}_{\phi}^k(c_t)=\\
\textrm{top}_{w\in \boldsymbol{x}}^k \phi\big (w; f(c_t), c_t\big) \textrm{ }\cup \textrm{ } \textrm{top}_{w\in \boldsymbol{y}_{<t}}^k \phi\big(w; f(c_t), c_t\big)
\end{multline*}
\noindent where $\cup$ denotes the union of two sets, and $\textrm{top}_{w\in \boldsymbol{x}}^k \phi(w; f(c_t), c_t)$ returns words corresponding to the $k$ largest $\phi$ values. ~\footnote{In fact, $\mathcal{W}_{\phi}^k(c_t) \rightarrow f(c_t)$
can be considered as generalized translation rules obtained by $\phi$. In other words, the rules are extracted under teacher forcing decoding. In particular, if $k=1$, this is similar to the statistical machine translation (SMT) with word level rules \cite{koehn2009statistical}, except that a generalized translation rule also involves a word from $\boldsymbol{y}_{<t}$ which simulates the role of language modeling in SMT.}

In addition, suppose $Q(y \mid \mathcal{W}_{\phi}^k(c_t); \theta)$ ($Q(\theta)$ or $Q$ for brevity) is a proxy model that makes a translation decision on top of $\mathcal{W}_{\phi}^k(c_t)$ rather than the entire context $c_t$ like a standard NMT model.
Formally, we define a principled metric as follows:
\begin{definition}
The metric of $\phi$ is defined by
\vspace{-10pt}
\begin{multline}
\min_{Q} \min_{\theta} -\mathbb{E}_{c_t} \Big[\log Q\big(f(c_t)\mid \mathcal{W}_{\phi}^k(c_t); \theta\big) \Big]
\label{eq:metric-org}
\end{multline}
\noindent where $\mathbb{E}_{c_t}[\cdot]$ denotes the expectation with respect to the data distribution of $c_t$, and $Q$ is minimized over all possible proxy models.
\end{definition}
The underlying idea of the above metric is to measure the expectation of the disagreement between an optimal proxy model $Q$ constructed from $\phi$ and the NMT model $P$. Here the disagreement is measured by the minus log-likelihood of $Q$ over the data $\langle \mathcal{W}_{\phi}^k(c_t), f(c_t) \rangle$ whose label $f(c_t)$ is generated from $P$.~\footnote{It is natural to extend our definition by using other similar disagreement measures such as the KL distance. Since the KL distance requires additional GPU memory to restore the distribution $P$ in the implementation, we employ the minus log-likelihood for efficiency in our experiments.}  

\paragraph{Definition of Fidelity}

The metric of $\phi$ actually defines fidelity by measuring how much the optimal proxy model defined on $\mathcal{W}^k_{\phi}(c_t)$ disagrees with $P(y \mid c_t)$.  The mention of \textit{fidelity} is widely used in model compression \cite{bucilua,DBLP:journals/corr/abs-1802-05668}, model distillation \cite{hinton2015distilling,liu-etal-2018-distilling}, and particularly in evaluating the explanation models for black-box neural networks \cite{lakkaraju2016interpretable,bastani2017interpreting}. 
These works focus on learning a specific model $Q$ on which fidelity can be directly defined. However, we are interested in evaluating explanation methods $\phi$ where $Q$ is a latent variable that we have to minimize. By doing this, fidelity in our metric is defined on $\phi$ as shown in Eq~\eqref{eq:metric-org}. 

\subsection{Approximation}
Generally, it is intractable to exactly calculate the principled metric due to two main challenges. 
On one hand, the real data distribution of $c_t$ is {unknowable}, making it impossible to exactly define the expectation with respect to an unknown distribution. 
On the other hand, the domain of a proxy model $Q$ is not bounded, and it is difficult to minimize a model $Q$ within an unbounded domain.

\paragraph{Empirical Risk Minimization}
Inspired by the statistical learning theory~\cite{vapnik1999overview}, we calculate the expected disagreement over $c_t$ by a two-step strategy:  we minimize the  empirical risk to obtain an optimized $\theta$ for a given $Q$;  and then we estimate the risk defined on a held-out test set by using the optimized $\theta$.
In this way, we cast the principled metric into a standard machine learning task.

For a given model architecture $Q$, to optimize $\theta$, we first collect the training set as $\{\langle \mathcal{W}_{\phi}^k(c_t), f(c_t)\rangle\}$  for each sentence pair $\langle\boldsymbol{x}, \boldsymbol{y}\rangle$ at every time step $t$, where $\langle\boldsymbol{x}, \boldsymbol{y}\rangle$  is a sentence pair from a given bilingual corpus $\mathcal{D}_{\textrm{train}} = \{\langle\boldsymbol{x}^n, \boldsymbol{y}^n \rangle\mid n=1, \cdots, N\}$.  Then we optimize $\theta$ by the empirical risk minimization:
\begin{equation}
\min_{\theta} \sum_{\langle\boldsymbol{x}, \boldsymbol{y}\rangle\in \mathcal{D}_{\textrm{train}}} \sum_{c_t} -\log Q(f(c_t) \mid \mathcal{W}_{\phi}^k(c_t); \theta) 
\label{eq:train}
\end{equation}

\paragraph{Proxy Model Selection}
In response to the second challenge of the unbounded domain, we define a surrogate distribution family ${\mathcal{Q}}$, and then approximately calculate Eq.\eqref{eq:metric-org} within ${\mathcal{Q}}$ instead:
\begin{equation}
 \min_{Q \in {\mathcal{Q}}} \min_{\theta} -\mathbb{E}_{c_t} \Big[\log Q\big(f(c_t)\mid \mathcal{W}_{\phi}^k(c_t); \theta\big) \Big]
 \label{eq:metric-appr}
 \end{equation}

We consider three different proxy models including multi-layer { feedforward network (FN)}, { recurrent network (RN)} and { self-attention network (SA)}. 
In details, for different networks $\epsilon \in \{\textrm{FN}, \textrm{RN}, \textrm{SA}\}$, the proxy model $Q^{\epsilon}$ is defined as follows:
\begin{equation*}
Q^{\epsilon}(y\mid \mathcal{W}_{\phi}^k(c_t)) = P(y\mid s_t^{\epsilon})
\end{equation*}
\noindent where $s_t^{\epsilon}$ is the decoding state regarding different architecture $\epsilon$. 
Specifically, for feedforward network, the decoding state is defined by $$s_t^{\textrm{FN}} = \textrm{FNN}(\tilde{x}_1, \cdots, \tilde{x}_k, \tilde{y}_1, \cdots, \tilde{y}_k).$$ {For $\epsilon \in\{ \textrm{RN}, \textrm{SA}\}$, the decoding state $s_t^{\epsilon}$ is defined by
$$s_t^{\epsilon} = \textrm{Attn}\big(s_0, \{ h_{\tilde{x}_1},\cdots, h_{\tilde{x}_k}, h_{\tilde{y}_1}  \cdots,h_{\tilde{y}_k} \}\big), $$
\noindent where $\tilde{x}$ and $\tilde{y}$ are source and target side words from $\mathcal{W}_{\phi}^k(c_t)$, $s_0$ is the query of init state, $h$ is the position-aware representations of words, generated by the encoder of \textrm{RN} or \textrm{SA} as defined in Eq.\eqref{eq:att} and Eq.\eqref{eq:tfm}.
For RN, $s_t^{\textrm{RN}}$ is the weight-sum vectors of a bidirectional LSTM over all selected top $k$ source and target words; while for SA, $s_t^{\textrm{SA}}$ is the weight-sum of vectors over the SA networks. }

\subsection{Evaluation Paradigm}

\begin{algorithm}
\caption{Calculating the evaluation metric}
\label{alg:eval}
\begin{algorithmic}[1]
\REQUIRE $\phi$, ${\mathcal{Q}}(\theta)$, $\mathcal{D}_{\textrm{train}}$, $\mathcal{D}_{\textrm{test}}$
\ENSURE the metric score $m$ of $\phi$ over $\mathcal{D}_{\textrm{test}}$ 
\STATE ${\mathcal{Q}}^*=\{\}$
\STATE Collect $\langle f(c_t), {\mathcal{W}_{\phi}^k(c_t)}\rangle$ from $\mathcal{D}_{\textrm{train}}$ and $\mathcal{D}_{\textrm{test}}$ to obtain two sets $\mathcal{FW_{\textrm{train}}}$ and $\mathcal{FW_{\textrm{test}}}$
\FOR{$Q(\theta) \in {\mathcal{Q}}(\theta)$}
\STATE Optimize $\theta^*$ over $\mathcal{FW_{\textrm{train}}}$ w.r.t Eq.\eqref{eq:train}
\STATE Add $Q(\theta^*)$ into ${\mathcal{Q}}^*$
\ENDFOR
\FOR{$Q^*\in{\mathcal{Q}}^*$}
\STATE{$m_{Q^*}=0$}
\FOR{$\langle f(c_t), \mathcal{W}_{\phi}^k(c_t)\rangle\in \mathcal{FW_{\textrm{test}}}$}
\STATE $m_{Q^*} \pluseq  -\log Q^*(f(c_t) \mid \mathcal{W}_{\phi}^k(c_t)) $
\ENDFOR
\ENDFOR
\STATE Return $\min\limits_{Q^*\in{\mathcal{Q}}^*}\exp\big(\frac{m_{Q^*}}{|\mathcal{FW_{\textrm{test}}}|}\big)$
\end{algorithmic}
\end{algorithm}

\noindent Given a bilingual training set $\mathcal{D}_{\textrm{train}}$ and a bilingual test set $\mathcal{D}_{\textrm{test}}$, we evaluate an explanation method $\phi$ w.r.t the NMT model $P(y\mid c_t)$ by setting the proxy model family ${\mathcal{Q}}(\theta)$ to include three neural networks as defined before. Following the standard process of addressing a machine learning problem, Algorithm~\ref{alg:eval} summarizes the procedure to approximately calculate the metric of $\phi$ on the test dataset $\mathcal{D}_{\textrm{test}}$, which returns the preplexity (PPL) on $\mathcal{FW_{\textrm{test}}}$.~\footnote{Note that the negative log-likelihood in Eq.~\ref{eq:metric-org} is proportional to PPL and thus we use PPL as the metric value in this paper.} 

In this paper, we try four different choices to specify the surrogate family, i.e., ${\mathcal{Q}}=\{Q^\textrm{FN}\}$, ${\mathcal{Q}}=\{Q^\textrm{RN}\}$, ${\mathcal{Q}}=\{Q^\textrm{SA}\}$, and ${\mathcal{Q}}=\{Q^\textrm{FN}, Q^\textrm{RN}, Q^\textrm{SA}\}$, leading to four instances of our metric respectively denoted as {\bf FN}, {\bf RN}, {\bf SA} and {\bf Comb}.
In addition, as the {\bf baseline} metric, we employ the well-trained NMT model $P$ as the proxy model $Q$ by masking out the input words that do not appear in the rule set $\mathcal{W}_{\phi}^k(c_t))$. 
{For the baseline metric, it doesn't require to train $Q's$ parameter $\theta$ and tests on $\mathcal{D}_{\textrm{test}}$ only.}
Since $P$ is trained with the entire context $c_t$ whereas it is testified on $\mathcal{W}_{\phi}^k(c_t)$, this mismatch may lead to poor performance and is thus less trusted. This baseline metric extends the idea of ~\citet{arras2016explaining,denil2014extraction} from classification tasks to structured prediction tasks like machine translation {which are highly dependent on context rather than just keywords. }

\section{Experiments}

In this section, we conduct experiments to prove the effectiveness of our metric from two viewpoints: how good an explanation method is and which explanation method is better than others.

\subsection{Settings}

\paragraph{Datasets} We carry out our experiments on three standard IWSLT translation tasks including IWSLT14 \DeEn (167k sentence pairs), IWSLT17 \ZhEn (237k sentence pairs) and IWSLT17 \FrEn (229k sentence pairs). All these datasets are tokenized and applied BPE (Byte-Pair Encoding) following ~\citet{ott2019fairseq}. The target side vocabulary sizes  of the three datasets are 8876, 11632, and 9844 respectively. In addition, we carry out extended experiments on three large-scale WMT translation tasks including WMT14 \DeEn (4.5m sentence pairs), WMT17 \ZhEn (22m sentence pairs) and WMT14 \FrEn (40.8m sentence pairs), with vocabulary sizes 22568, 29832, 27168 respectively.

\paragraph{NMT Systems} To examine the generality of our evaluation method, we conduct experiments on two NMT systems, i.e. \textsc{Rnn-Search} (denoted by {\bf RNN}) and \textsc{Transformer} (denoted by {\bf Trans.}), both of which are implemented with fairseq~\cite{ott2019fairseq}. For RNN, we adopt the 1-layer RNN with LSTM cells whose encoder (bi-directional) and decoder hidden units are 256 and 512 respectively. For \textsc{Transformer} on the IWSLT datasets, the number of layers and attention heads are 2 and 4 respectively. For both models, we set the embedding dimensions as 256. On WMT datasets, we simply use \textsc{Transformer-base} with 4 attention heads.
The performances of our NMT models are comparable to those reported in recent literature \cite{Tan2019MultilingualNM}.

\paragraph{Explanation Methods} On both NMT systems, we implement four explanation methods, i.e. Attention (\textsc{Attn}), gradient norm (\textsc{Ngrad}), weighted gradient (\textsc{Wgrad}), and prediction difference (\textsc{Pd}) as mentioned in Section \S 2.

\paragraph{Our metric} We implemented five instantiations of the proposed metric including FN, RN, SA, Comb, and Baseline (Base for brevity) as presented in section \S3.3. To configurate them, we adopt the same settings from NMT systems to train SA and RN. FN is implemented with feeding the features of bag of words through a 3-layer fully connected network. 
{As given in algorithm \ref{alg:eval}, the approximate fidelity is estimated through $Q$ with the lowest PPL, therefore the best metric is that achieves the lowest PPL since it results in a closer approximation to the real fidelity.}

\subsection{Experiments on IWSLT tasks}
In this subsection, we first {conduct} experiments and analysis on the IWSLT \DeEn task to configurate fidelity-based metric and then extend the experiments to other IWSLT tasks. 

\paragraph{Comparison of metric instantiations}

\begin{table}[t]
\begin{center}
\setlength{\tabcolsep}{1.5pt}
\scalebox{0.86}{
\begin{tabular}{c||c||cccc}
\textbf{NMT} & \textbf{Metric} & \textsc{Attn} & \textsc{Pd} & \textsc{Ngrad} &\textsc{Wgrad} \\ \hline\hline
\multirow{5}{*}{\textbf{Trans}} & \textbf{Base} & 196.9 &54.3 &193.4 &13400\\ 
&  \textbf{ FN} & 13.9 & 5.8 & 11.3 & 131.2\\
&  \textbf{ RN} & 13.8 & 5.7 & 10.7 & 126.7\\
&  \textbf{ SA} & 13.9 & 5.5 & 10.8 & 119.5\\
&  \textbf{Comb} & 13.8 & 5.5 & 10.7 & 119.5\\\hline\hline

\multirow{5}{*}{\textbf{RNN}} & \textbf{Base} & - &54.2 &90.3 &28587\\ 
&  \textbf{ FN} & - & 6.7 & 8.3 & 170.8\\
&  \textbf{ RN} & - & 6.5 & 7.8 & 163.2\\
&  \textbf{ SA} & - & 6.5 & 8.1 & 154.9\\
&  \textbf{Comb} & - & 6.5 & 7.8 & 154.9\\
\end{tabular}
}
\end{center}
\caption{\label{tab:all-metric-results}The PPL comparison for the five metric instantiations on the IWSLT \DeEn dataset. } 
\end{table}

We calculate PPL on the IWSLT \DeEn dataset for four metric instantiations (FN, RN, SA, Comb) and Baseline (Base) with $k=1$ to extract the most relevant words. 
Table~\ref{tab:all-metric-results} summarizes the results for two translation systems (\textsc{Transformer} annotated as {Trans} and \textsc{Rnn-Search} annotated as {RNN}), respectively. 
Note that since there is no target-side attention in \textsc{Rnn-Search}, we can not extract the best relevant target word, so Table~\ref{tab:all-metric-results} does not include the results of \textsc{Attn} method for \textsc{Rnn-Search}.

The baseline (Base) achieves undesirable PPL which indicates the relevant words identified by \textsc{Pd} failed to make the same decision as the NMT system. The main reason is that the mismatch between training and testing leads to the issue as presented in section \S3.3.
On the contrary, the other four metric instantiations attain much lower PPL than the Baseline. In addition, the PPLs on \textsc{Pd}, \textsc{Ngrad}, and \textsc{Attn} are much better than those on \textsc{Wgrad}.
This finding shows that all \textsc{Pd}, \textsc{Ngrad}, and \textsc{Attn} are good explanation methods except \textsc{Wgrad} in terms of fidelity.

\paragraph{Density of generalizable rules}

To understand possible reasons for why one explanation method is better under our metric, we make a naive conjecture: when it tries to reveal the patterns that the well-trained NMT has captured, it extracted more concentrated patterns. In other words, a generalized rule $\mathcal{W}_{\phi}^k(c_t)\rightarrow f(c_t)$ from one sentence pair can often be observed among other examples.
 
To measure the density of the extracted rules, we first divide all extracted rules into five bins according to their frequencies. Then we collect the number of rules in each bin as well as the total number of rules. 
Table~\ref{tab:density} shows the statistics to measure the density of rules obtained from different evaluation methods. From this table, we can see that the density for \textsc{Pd} is the highest among those for all explanation methods, because it contains fewer infrequent rules in $B_1$, whereas there are more frequent rules in other bins. This might be one possible reason that \textsc{Pd} is better under our fidelity-based evaluation metric.
\begin{table}[t]
\begin{center}
\scalebox{0.86}{
\setlength{\tabcolsep}{2.pt}
\begin{tabular}{c||c|ccccc}
{\textbf{Method}}          &  \textbf{Total}  & $B_1$ & $B_2$& $B_3$& $B_4$ &$B_5$ 
      \\ \hline\hline
    \textsc{Attn}     &   1.97M & 1.65M  & 298K  & 23.7K & 1.54K &  104    \\
    \textsc{Pd} &   1.62M & 1.25M  & 328K  & 31.2K & 2.11K & 108   \\
    \textsc{Ngrad}    &   1.89M & 1.54M  & 326K  & 27.6K & 1.64K &  83 \\     
    \textsc{Wgrad}    &   2.62M & 2.37M  & 278K  & 17.5K & 0.86K &  34  \\
\end{tabular}
}
\caption{\label{tab:density}Density of the extracted rules from \textsc{Transformer} on the IWSLT \DeEn. The density is measured by the total number of unique rules and the number of rules with certain frequency in each interval $B_i$: $B_1=(0,1]$, $B_2=(1,10]$, $B_3=(10, 100]$, $B_4=(100, 1000]$, and $B_4=(1000,\infty$).}
\end{center}
\end{table}

\paragraph{Stability of ranking order} 
In Table~\ref{tab:all-metric-results} the ranking order is \textsc{Pd} $>$ \textsc{Ngrad} $>$ \textsc{Attn} $>$ \textsc{Wgrad} regarding all five metric instantiations. Generally, a good metric should preserve the ranking order of explanation methods independent of the test dataset. Regarding this criterion of order-preserving property, we analyze the stability of different fidelity-based metric instantiations. To this end, we randomly sample one thousand test data with replacement whose sizes are variant from 1\% to 100\% and then calculate the rate whether the ranking order is preserved on these test datasets. The results in Table~\ref{tab:stability-results} indicate that FN, RN, SA, Comb are more stable than Base to the change of distribution of test sets.

According to Table~\ref{tab:all-metric-results} and Table~\ref{tab:stability-results}, SA performs similar to the best metric Comb and it is faster than Comb or RN for training and testing, thereby, in the rest of experiments, we mainly employ SA to measure evaluation methods.

\begin{table}[t]
\begin{center}
\scalebox{0.85}{
\setlength{\tabcolsep}{3.3pt}
\begin{tabular}{c|ccccc}
       & Base & FN & SA & RN & Comb  \\ \hline\hline
    \textbf{1\%}   & 53.0\%  &  97.1\% & 99.9\% & 99.8\% & 99.8\%  \\
    \textbf{5\%}   & 56.1\%  &  100\%  & 100\%  & 100\%  & 100\%   \\
    \textbf{20\%}  & 60.8\%  &  100\%  & 100\%  & 100\%  & 100\%   \\
    \textbf{50\%}  & 66.8\%  & 100\%   & 100\%  & 100\%  & 100\%   \\
    \textbf{100\%} & 75.4\%  & 100\%   & 100\%  & 100\%  & 100\%   \\
\end{tabular}
}
\caption{\label{tab:stability-results}The rate (percentage) of sampled test dataset that have the same rankings as the test set on the IWSLT \ZhEn dataset.}
\end{center}
\end{table}

\paragraph{Effects on different $k$}

\begin{figure}
	\includegraphics[width=0.42\textwidth]{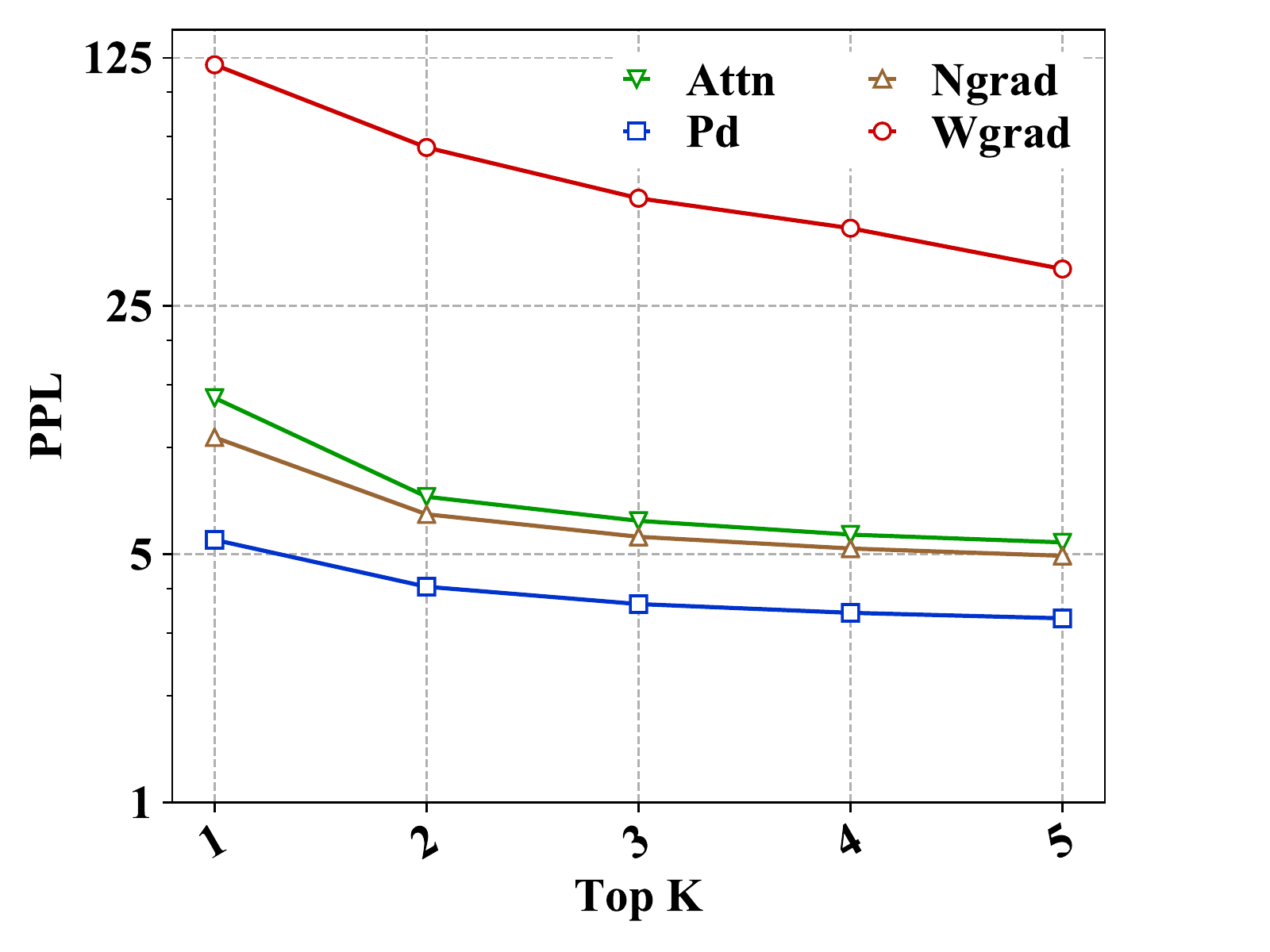}
	\caption{PPL for each explanation method on \textsc{Transformer} over the IWSLT \DeEn dataset with different $k$ value.}
	\label{fig:k}
\end{figure}

In this experiment, we examine the effects of explanation methods on larger $k$ with respect to SA. Figure~\ref{fig:k} depicts the effects of $k$ for \textsc{Transformer} on \DeEn task. One can clearly observe two findings: 1) the ranking order of explanation methods is invariant for different $k$.
2) as $k$ is larger, the PPL is much better for each explanation method. 3) the PPL improvement for \textsc{Pd}, \textsc{Attn}, and \textsc{Ngrad} is less after $k>2$, which further validates that they are powerful in explaining NMT using only a few words.  

\begin{table}[t]
\begin{center}
\scalebox{0.86}{
\setlength{\tabcolsep}{1.5pt}
\begin{tabular}{c|c||cccc}
\multirow{2}{*}{\textbf{NMT }} & \multirow{2}{*}{\textbf{Methods}} & \multicolumn{2}{c}{\textbf{ \ZhEn}} & \multicolumn{2}{c}{\textbf{ \FrEn}} \\
& & \textbf{Base} & \textbf{SA} & \textbf{Base} & \textbf{SA} \\\hline\hline

\multirow{4}{*}{\textbf{{Trans}}} & \textsc{Attn} & 897.1 &30.8 &359.6 &12.1\\
&  \textsc{Pd} & 215.1 & 10.8 & 55.3 & 4.6\\
&  \textsc{Ngrad} & 583.7 & 19 & 271.0 & 8.7\\
&  \textsc{Wgrad} & 24126 & 180.9 & 44287 & 155.4\\\hline

\multirow{4}{*}{\textbf{{RNN}}} & \textsc{Attn} & - & - & - & -\\ 
&  \textsc{Pd} & 139.9 & 11.3 & 49.0 & 5.5\\
&  \textsc{Ngrad} & 263.0& 13.2 & 85.8 & 6.7\\
&  \textsc{Wgrad} & 23068 & 243.1 & 50657 & 194.9\\

\end{tabular}
}

\end{center}
\caption{\label{tab:iwslt-zhfr}The PPL comparison for two fidelity-based metric instantiations on two IWSLT datasets. } 
\end{table}

\paragraph{Testing on other scenarios}
In the previous experiments, our metric instantiations are trained and evaluated under the same scenario, where $c_t$ used to extract relevant words is obtained from gold data and its label $f(c_t)$ is the prediction from NMT $f${, namely Teacher Forcing Decode}. 
To examine the robustness of our metric, we apply the trained metric to two different scenarios: real decoding scenario (Real-Decode) where both $c_t$ and its label $f(c_t)$ are from the NMT output; and golden data scenario (Golden-Data) where both $c_t$ and its label are from golden test data. The results for both scenarios are shown in Table~\ref{tab:scenarios}.

From Table~\ref{tab:scenarios}, we see that the ranking order for both scenarios is the same as before. To our surprise, the results in Real-Decode are even better than those in the matched Teacher Forcing Decode scenario. One possible reason is that the labels generated by a NMT system in the Real-Decode tend to be high-frequency words, which leads to better PPL.
In contrast, our metric instantiation in the Golden-Data results in much higher PPL due to the mismatch between training and testing. 
The performance of experimenting training and testing in the same scenario like Golden-Data can be experimented in future works, however, it's not the focus of this paper.

\begin{table}[t]
\begin{center}
\scalebox{0.86}{
\begin{tabular}{c||ccc}
\textbf{Methods} &\textbf{R-Dec}  &  \textbf{Golden} & \textbf{T-Dec} \\\hline\hline
\textsc{Attn}   &  11.5  &  57.1   & 13.8\\
\textsc{Pd}     &  4.7  &  23.3   & 5.5\\
\textsc{Ngrad}   & 8.2   & 42.0  & 10.7\\
\textsc{Wgrad}  & 115.0  & 223.4   & 119.5\\ 
\end{tabular}
}
\end{center}
\caption{\label{tab:scenarios}Evaluating four explanation methods on 3 different scenarios Real-Decode (R-Dec), Golden-Data (Golden) and Teacher-Forcing Decode (T-Dec)) for \textsc{Transformer} over IWSLT \DeEn task. }
\end{table}

\subsection{Scalability on WMT tasks}

Since our metric such as SA requires to extract generalized rules for each explanation method from the entire training dataset, it is computationally expensive for some explanation methods such as gradient methods to directly run on WMT tasks with large scale training data.

\begin{figure}
	\includegraphics[width=0.42\textwidth]{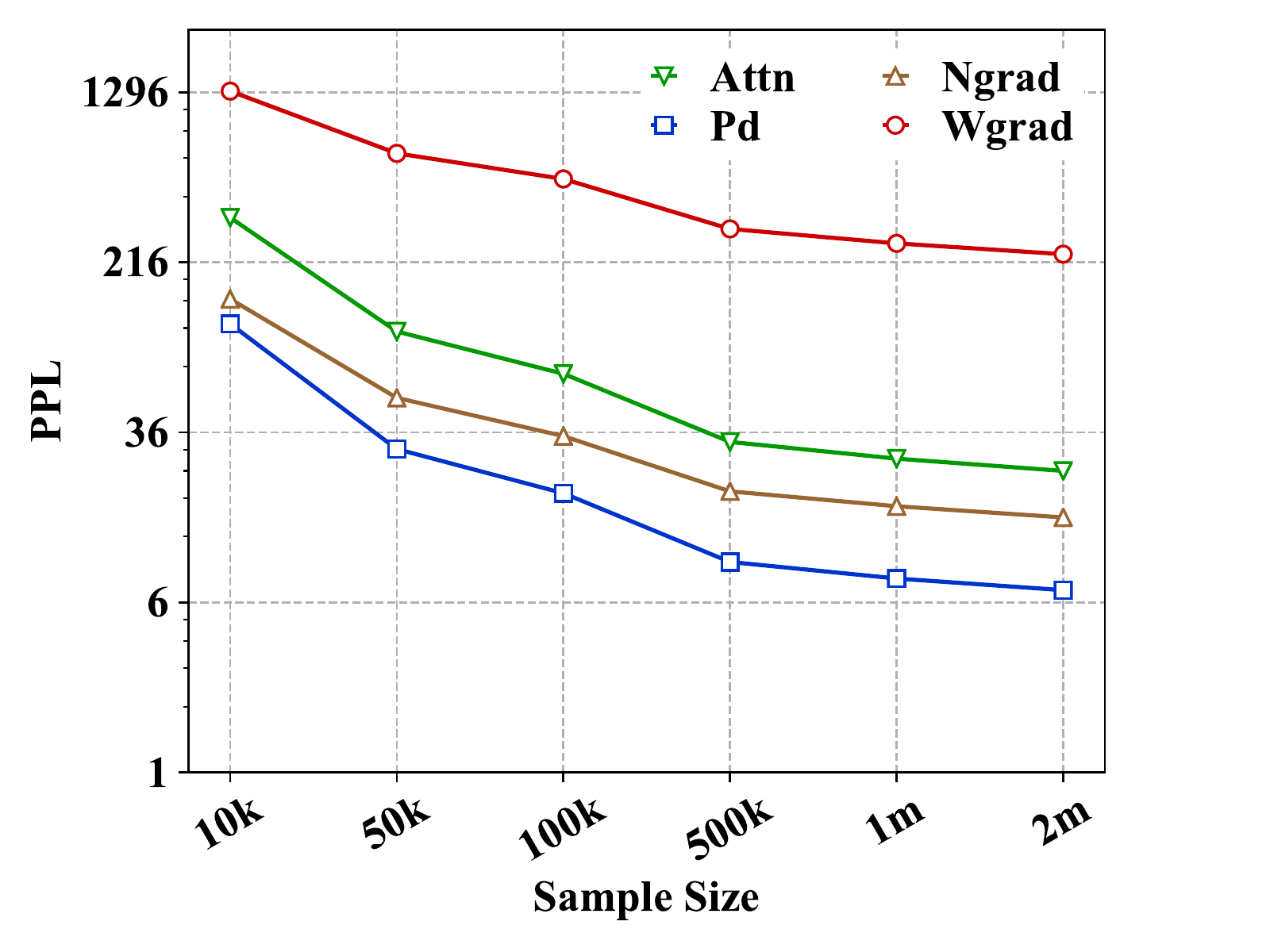}
	\caption{PPL for each explanation method on \textsc{Transformer} over WMT \ZhEn task with different sample sizes. }
	\label{fig:sample}
\end{figure}

\paragraph{Effects on sample size}
We randomly sample some subsets over WMT \ZhEn training data that includes 22 million sentence pairs to form several new training sets. The sample sizes of the new training sets are set up to 2 million and the results are illustrated in Figure~\ref{fig:sample}. The following facts are revealed. Firstly, the ranking order of four explanation methods remains unchanged with respect to different sample sizes. Secondly, with the increase of the sample size, the metric score decreases slower and slower and there is no significant drop from sampling 2 million sentence pairs to sampling 1 million.

\begin{table}[t]
\begin{center}
\scalebox{0.86}{
\setlength{\tabcolsep}{1.5pt}
\begin{tabular}{c|c||cccc}
\multirow{2}{*}{\textbf{Datasets }} & \multirow{2}{*}{\textbf{Methods}} & \multicolumn{2}{c}{\textbf{Base}} & \multicolumn{2}{c}{\textbf{ SA}} \\
& & \textbf{PPL} & \textbf{Rank} & \textbf{PPL} & \textbf{Rank} \\\hline\hline

\multirow{4}{*}{\textbf{ \ZhEn}} & \textsc{Attn} & 336.4 &\underbar{2} &27.3 &\underbar{3}\\
&  \textsc{Pd} &165.3 &1 & 7.7 &1\\
&  \textsc{Ngrad} &435.2 &\underbar{3} & 16.5 &\underbar{2}\\
&  \textsc{Wgrad} &1615.5 &4 & 263.5 &4\\\hline

\multirow{4}{*}{\textbf{ \DeEn}} & \textsc{Attn} & 1862.3 &\underbar{2} &17.0 &\underbar{3}\\
&  \textsc{Pd} &1118.2 &1 & 5.4 &1\\
&  \textsc{Ngrad} &2827.7 &\underbar{3} & 15.1 &\underbar{2}\\
&  \textsc{Wgrad} &6678.1 &4 & 197.4 &4\\\hline

\multirow{4}{*}{\textbf{ \FrEn}} & \textsc{Attn} & 4271.0 &3 &41.1 &3\\
&  \textsc{Pd} &1646.6 &1 & 4.1 &1\\
&  \textsc{Ngrad} &2810.2 &2 & 11.8 &2\\
&  \textsc{Wgrad} &6703.8 &4 & 163.7 &4\\
\end{tabular}
}

\end{center}
\caption{\label{tab:scalability}The PPL and Ranking Order comparison between two fidelity-based metric instantiations (Base and SA) on three WMT datasets. `` \underbar{ } " denotes the mismatch of ranking order.} 
\end{table}

\paragraph{Results on WMT}
With the analysis of effects on various sample sizes, we choose a sample size of 1 million for the following scaling experiments. The PPL results for WMT \DeEn, \ZhEn ,and \FrEn are listed in Table~\ref{tab:scalability}. 
We can see that the order \textsc{Pd} $>$ \textsc{Ngrad} $>$ \textsc{Attn} $>$ \textsc{Wgrad} evaluated by SA still remains unchanged on these three datasets as before. 
One can observe that the ranking order under the baseline doesn’t agree with SA on WMT \DeEn and \ZhEn. Since the baseline yields in high PPL due to the mismatch we mentioned in section \S 3.3 ,in this case, we tend to trust the evaluation results from SA that achieves lower PPL leading to better fidelity.

\subsection{Relation to Alignment Error Rate}

\begin{table}[t]
\begin{center}
\scalebox{.85}{
\setlength{\tabcolsep}{1.pt}
\begin{tabular}{c|c||cccc}
\multirow{2}{*}{\textbf{Datasets}} & \multirow{2}{*}{\textbf{Methods}} & \multicolumn{2}{c}{\textbf{SA}} & \multicolumn{2}{c}{\textbf{Alignment}} \\
& & \textbf{PPL} & \textbf{Rank} & \textbf{AER} & \textbf{Rank} \\\hline\hline

\multirow{4}{*}{\textbf{IWSLT \ZhEn}} & \textsc{Attn} & 30.8 & 3 &55.0 &3\\
&  \textsc{Pd} &10.8 &1 & 50.6 &1\\
&  \textsc{Ngrad} &19 & 2 & 52.9 &2\\
&  \textsc{Wgrad} &180.9 &4 & 79.2 &4\\\hline

\multirow{4}{*}{\textbf{WMT \ZhEn}} & \textsc{Attn} & 27.3 &\underbar{3} &42.1 &\underbar{2}\\
&  \textsc{Pd} &7.7 &1 & 32.7 &1\\
&  \textsc{Ngrad} &16.5 &\underbar{2} & 49.3 &\underbar{3}\\
&  \textsc{Wgrad} &263.5 &4 & 79.2 &4\\\hline
\multirow{4}{*}{\textbf{WMT \DeEn}} & \textsc{Attn} & 17.0 &3 &48.7 &3\\
&  \textsc{Pd} &5.4 &1 & 34.1 &1\\
&  \textsc{Ngrad} &15.1 &2 & 48.1 &2\\
&  \textsc{Wgrad} &194.7 &4 & 73.5 &4\\

\end{tabular}
}

\end{center}
\caption{\label{tab:aer-cmp}Relation with word alignment. `` \underbar{ } " denotes the mismatch of ranking order.} 
\end{table}

\begin{figure}
	\includegraphics[width=0.475\textwidth]{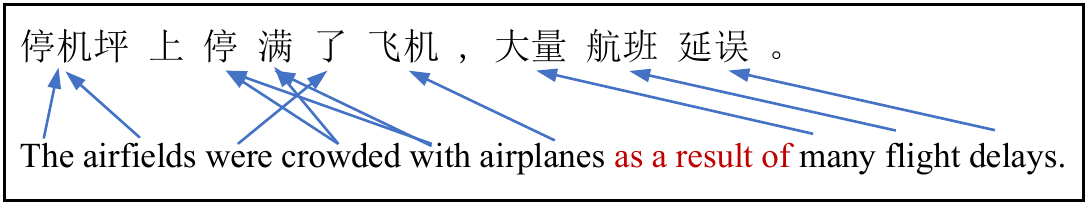}
	\caption{AER can not evaluate explanation methods on those target words ``as a result of", which are not aligned to any word in the source sentence according to human annotation.}
	\label{fig:case}
\end{figure}

Since the calculation of the Alignment Error Rate (AER) requires manually annotated test datasets with ground-truth word alignments, we select three different test datasets contained such alignments for experiments, namely, IWSLT \ZhEn, NIST05 \ZhEn \footnote{\url{https://www.ldc.upenn.edu/collaborations/evaluations/nist}} and Zenkel \DeEn~ \cite{zenkel2019adding}. Note that unaligned target words account for 7.8\%, 4.7\%, and 9.2\% on these three test sets respectively, which are skipped by AER for evaluating explanation methods. 
For example, in Figure~\ref{fig:case}, those target
words `as a result` cannot
be covered by AER due to the impossibility of human annotation,
but for a fidelity-based metric, they can be analyzed as well.

Table \ref{tab:aer-cmp} demonstrates that our fidelity-based metric does not agree very well with AER on the WMT \ZhEn task:
 \textsc{Ngrad} is better than  \textsc{Attn} in terms of SA but the result is opposite in terms of AER.
 Since the evaluation criteria of SA and AER are different, it is reasonable that their evaluation results are different.
 This finding is in line with the standpoint by~\citet{Jacovi2020TowardsFI}: SA is an objective metric that reflects fidelity of models while AER is a subject metric based on human evaluation. However, it is observed that the ranking by SA is consistent on all three tasks but that by AER is highly dependent on different tasks.

\section{Related Work}

In recent years, explaining deep neural models has been a growing interest in the deep learning community, aiming at more comprehensible and trustworthy neural models. In this section, we mainly discuss two dominating ways towards it. One way is to develop explanation methods to interpret a target black-box neural network~\cite{bach2015pixel,zintgraf2017visualizing}. 
For example, on classification tasks, ~\citet{bach2015pixel} propose layer-wise relevance propagation to visualize the relationship between a pair of neurons within networks, and \citet{li2015visualizing} introduce a gradient-based approach to understanding the compositionality in neural networks for NLP.  In particular, on structured prediction tasks, many research works design similar methods to understand NMT models~\cite{ding2017visualizing,alvarez2017causal,koehn2019saliency,he-etal-2019-towards}. 

The other way is to construct an interpretable model for the target network and  then indirectly interpret its behavior to understand the target network on classification tasks~\cite{lei2016rationalizing, murdoch2017automatic, arras2017relevant,wang2019multi}. The interpretable model is defined on top of extracted rational evidence and learned by model distillation from the target network.
To extract rational evidence from the entire inputs, one either leverages a particular explanation method~\cite{lei2016rationalizing,wang2019multi} 
or an auxiliary evidence extraction model~\cite{murdoch2017automatic,arras2017relevant}. Although our work focuses on evaluating explanation methods and does not aim to construct an interpretable model, we draw inspiration from their ideas to design $Q\in \mathcal{Q}$ in Eq.~\eqref{eq:metric-org} for our evaluation metric. 

With the increasing efforts on designing new explanation methods, yet there are only a few works proposed to evaluate them. ~\citet{mohseni2018human} propose a paradigm to evaluate explanation methods for document classification that involves human judgment for evaluation. ~\citet{poerner-etal-2018-evaluating} conduct the first human-independent comprehensive evaluation of explanation methods for NLP tasks. However, their metrics are task-specific because they make some assumptions for a specific task. Our work proposes a principled metric to evaluate explanation methods for NMT and our evaluation paradigm is independent of any assumptions as well as humans. It is worth noting that~\citet{arras2016explaining,denil2014extraction} directly measure the performance of the target model $P$ on the extracted words without constructing $Q$ to evaluate explanation methods for classification tasks. However, since translation is more complex than classification tasks, $P$ trained on the entire context $c_t$ typically makes a terrible prediction when testing on the compressed context $\mathcal{W}_{\phi}^k(c_t)$. As a result, the poor prediction performance makes it difficult to discriminate one explanation method from others, as observed in our internal experiments.
Concurrently, \citet{Jacovi2020TowardsFI} make a proposition to evaluate faithfulness of an explanation method separately from readability and plausibility (i.e., human-interpretability), which is similar to our definition of fidelity, but they do not formalize a metric or propose algorithms to measure it.

\section{Conclusions}
This paper has made an initial attempt to evaluate explanation methods from a new viewpoint. 
It has presented a principled metric based on fidelity in regard to the predictive behavior of the NMT model. 
Since it is intractable to exactly calculate the principled metric for a given explanation method, it thereby proposes an approximate approach to address the minimization problem. The proposed approach does not rely on human annotation and can be used to evaluate explanation methods on all target words. 
On six standard translation tasks, the metric quantitatively evaluates and compares four different explanation methods for two popular translation models. Experiments reveal that \textsc{Pd}, \textsc{Ngrad}, and \textsc{Attn} are all good explanation methods that are able to construct the NMT model's predictions with relatively low perplexity and \textsc{Pd} shows the best fidelity among them.

\section*{Acknowledgments}
We would like to thank all anonymous reviews for their valuable suggestions. This research was supported by Tencent AI Lab.

\bibliography{acl2020}
\bibliographystyle{acl_natbib}

\end{document}